\pdfoutput=1
\documentclass[11pt]{article}

\usepackage[final]{acl}

\usepackage{times}
\usepackage{latexsym}

\usepackage[T1]{fontenc}
\usepackage[utf8]{inputenc}

\usepackage{times}
\usepackage{latexsym}
\usepackage{booktabs}
\usepackage{amsmath}
\usepackage{float}
\usepackage{listings}
\usepackage{adjustbox}
\usepackage{longtable}
\usepackage{xcolor,colortbl}
\usepackage{diagbox} % added
\usepackage{array,multirow}
\usepackage{graphicx} % added
\usepackage{placeins} % added
\usepackage{microtype} % added
\usepackage{enumitem} % added
\usepackage{amsmath, bm} % added
\usepackage{comment} % added
\usepackage{amsmath, bm, amssymb}
\usepackage{subcaption}
\usepackage{transparent}
\usepackage{geometry}
\usepackage{pdflscape}
\usepackage{tcolorbox}
\tcbuselibrary{listings, breakable}
\definecolor{grey}{named}{gray}
\usepackage{listings}
\usepackage{hyperref}
\usepackage{url}
\usepackage{xurl}

\usepackage{microtype}

\usepackage{inconsolata}

\definecolor{mF}{HTML}{A9D18E} % Green
\definecolor{mNDCG}{HTML}{5187ED} % Blue
\definecolor{mQ}{HTML}{FFD966} % Yellow
\newcommand{\ccell}[3]{%
    \cellcolor{#1!\fpeval{#2*100}}#3%
}

\title{How Important is Recall for Measuring Retrieval Quality?}

\author{Shelly Schwartz\thanks{Work done while at Primer Technologies.}, Oleg Vasilyev\footnotemark[1], Randy Sawaya \\
  Primer Technologies Inc. \\
  San Francisco, California \\
  \texttt{{shellyganga@gmail.com, vasilyev.oleg@gmail, randy.sawaya}@primer.ai}\\}

\begin{document}
\maketitle
\begin{abstract}
In realistic retrieval settings with large and evolving knowledge bases, the total number of documents relevant to a query is typically unknown, and recall cannot be computed. In this paper, we evaluate several established strategies for handling this limitation by measuring the correlation between retrieval quality metrics and LLM-based judgments of response quality, where responses are generated from the retrieved documents. We conduct experiments across multiple datasets with a relatively low number of relevant documents (2–15). We also introduce a simple retrieval quality measure that performs well without requiring knowledge of the total number of relevant documents.
\end{abstract}

\section{Introduction}\label{sec:Introduction}
As an important and well-known example of information retrieval, Retrieval-Augmented Generation(RAG) involves gathering a set of query-relevant documents (or chunks of documents) from a knowledge base (KB), then sending the highest ranked $K$ documents to a Large Language Model (LLM) in order to generate a response to a particular query.  In an ideal scenario, the majority of the top $K$ documents would be relevant to the query in addition to making up a significant amount of the total relevant documents in the KB.

A common measure of the quality of the top $K$ selection is the $F$ measure, a harmonic weighted average of precision and recall \cite{Rijsbergen1977ATB, Rijsbergen1979}. In a real, changing KB the total number of relevant documents is not known in advance for an arbitrary query, meaning that the recall is also not known \cite{webber2012approximate, yilmaz2006estimating, kutlu2017intelligent, scholer2006capturing, fan2022nearly, ferrante2021meaningful, soboroff2006comparison, nguyen2018active, zobel1998reliability}.

There are several ways to circumvent the lack of knowledge required to calculate recall:
\begin{enumerate}[topsep=0pt,itemsep=-1ex,partopsep=1ex,parsep=1ex]
    \item Creating a fixed benchmark: Given a set of queries, a corresponding set of relevant documents is known and makes up a subset of the KB.  The subset is fixed and must be recreated when substantial changes are made to the KB.
    \item Estimating the total number of relevant documents for each query: This can be done by (1) attempting to extract as many relevant documents as possible (known as pooling) \cite{upadhyay2024llmspatch, abbasiantaeb2024llmfill, 10.1145/3726302.3729916, ganguly2023queryspecificvariabledepthpooling}, (2) extrapolating the count of relevant documents to higher $K$ \cite{rathee2025breakinglenstelescopeonline, schmidt2025effectivedatapruningscore, krishna-etal-2025-fact} or (3) comparing two different retrieval methods (known as capture–recapture) \cite{fraysse2023populationsizeestimationcapturerecapture, mordido2020markevaluate}.
    \item Using an evaluation measure that does not include the total number of relevant documents.
\end{enumerate}

We investigate the relative quality of a few simple measures which utilize these approaches. For example, $nDCG$ \cite{10.1145/345508.345545, 10.1145/582415.582418} does not use the total number of relevant (``positive'') documents $N_p$. In~\cite{10.1007/s10791-020-09377-x} $nDCG$ was noted for high discriminative power for top-$K$ recommendations and, in~\cite{10.1145/1835449.1835560}, for having strong correlations with interleaving.

\begin{figure*}[t!]
    \centering
    \includegraphics[width=\textwidth]{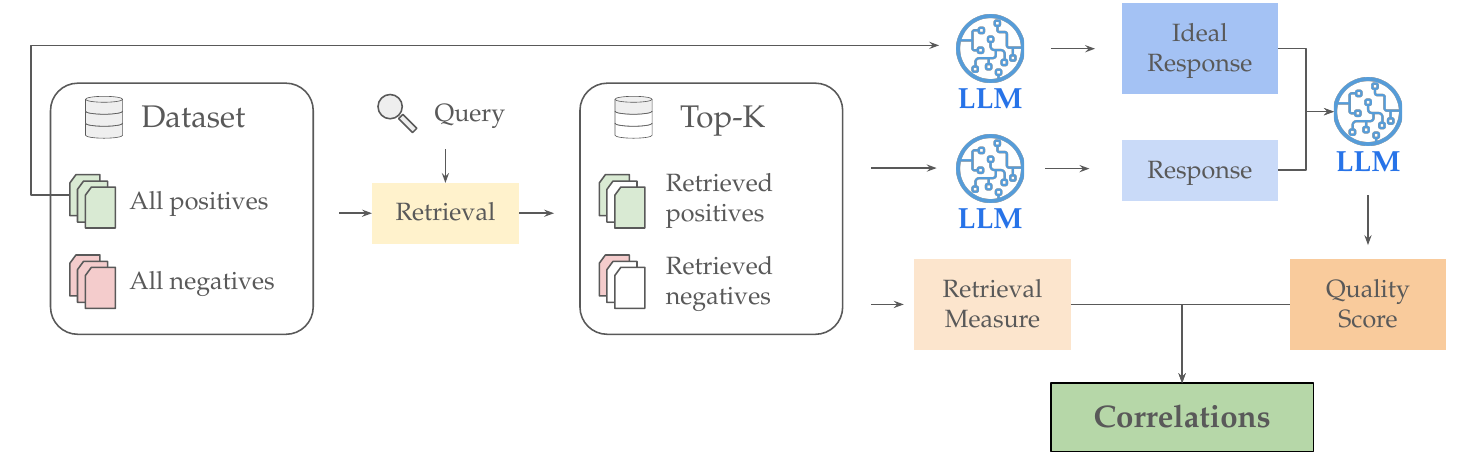}
    \caption{Obtaining a correlation between a measure of quality of retrieval and LLM judgment score.}
    \label{fig:schema_all}
\end{figure*}

We use several datasets with a low $N_p$ in $[2,15]$ and $K$ covering a range around $N_p$ as well as a few common, small embeddings for retrieval. We judge the measures by their correlations with the quality of LLM response to the query as both the measures and the response use the top $K$ retrieved documents. This is illustrated in Figure~\ref{fig:schema_all} and explained in the next section.

% See figure in: https://docs.google.com/drawings/d/1aSNUhYULI6hPJL-Kx-LP5dq1r6WZEpajR0ws7oB7hg4

In related work, the recall was noted as more important than precision~\cite{li2025doesknowledgeselectionhelp}. It is also known that the irrelevant documents in the top-$K$ do cause problems~\cite{amiraz-etal-2025-distracting, bdcc8090115} and the order of the top-$K$ texts can be important~\cite{ma2025inputordershapesllm, guo-etal-2024-makes, Xia_Zhou_Shi_Chen_Huang_2025}.

Our contribution:
\begin{enumerate}[topsep=0pt,itemsep=-1ex,partopsep=1ex,parsep=1ex]
    \item We introduce a simple measure ``$T$'' (as in \textbf{T}op-$K$ selection) independent of the total number $N_p$ of KB documents relevant to the query; we argue the measure is well founded and convenient for real-life evaluations.
    \item Taking a diverse selection from well established datasets, we compose a dataset consisting of queries, positives, negatives, LLM-generated answers from the retrieved results and LLM-generated scores of quality of the generated results. The retrieval is done by several popular embedding models.
    \item Using the above dataset, we obtain and review the correlations of $F$, $F_e$ ($F$ with a simple estimate of $N_p$), $nDCG$ and $T$ measures with the LLM quality score. We gather insights on the relative weaknesses and strengths of each of these measures.
\end{enumerate}

\section{Setup}\label{sec:Setup}
\subsection{Measures}\label{ssec:Measures}
As a benchmark measure we use the well known $F$ measure~\cite{Rijsbergen1979}
\begin{equation} \label{eq:Fa_RP}
F = \frac{1}{\frac{\alpha}{P} + \frac{1-\alpha}{R}}
\end{equation}
with a parameter $\alpha \in [0, 1]$. Here $P=\frac{n_p}{K}$ is the precision and $R=\frac{n_p}{N_p}$, the recall where $n_p$ represents the number of positives (relevant documents) within the top $K$ selection. This is equivalent to
\begin{equation} \label{eq:F_alpha_KNp}
F = \frac{n_p}{\alpha K + (1-\alpha)N_p}
\end{equation}
More commonly $F$ is used with the parametrization $\beta^2=\frac{1}{\alpha} - 1$.

We also consider ``$F_e$'' ($F$-estimated) - equivalent to $F$, but with $N_p$ restricted to the number of positives in the top 2$K$ rather than the total number of positives. This can be considered as a simple estimate of $N_p$, especially useful for low $N_p$, when an estimate by an exact extrapolation to large $K$ or by a capture-recapture method may be too crude.

\begin{figure*}[t!]
    \centering
    \includegraphics[width=\textwidth]{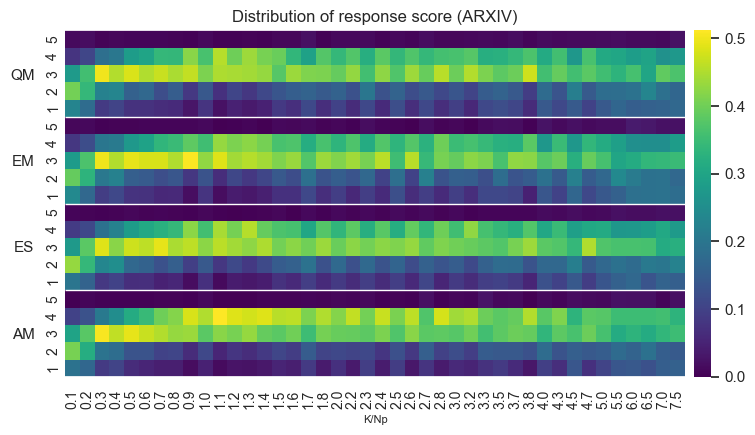}
    \caption{Distribution of the response score (1 to 5) for embedding models shown on Y-axis. On ARXIV; using only segments with minimum 300 samples, the ratio $\frac{K}{N_p}$ is rounded to the first digit.}
    \label{fig:Grade_DatA}
\end{figure*}

For the case of not knowing (or estimating) $N_p$, we consider $nDCG$~\cite{10.1145/345508.345545, 10.1145/582415.582418, wang2013theoreticalanalysisndcgtype}. 
We also suggest a simple measure
\begin{equation} \label{eq:T}
T = (1-\alpha)n_p - \alpha \frac{n_n}{K}
\end{equation}
where $n_n$=$K$-$n_p$ is the number of negative (irrelevant) documents within $K$. Similarly to $F$, $T$ includes a dependency on the parameter $\alpha$. For $F$, the parameter $\alpha$ may be selected for a desired weighted tradeoff between the precision and recall; for $T$ it gives a tradeoff between importance of positives and negatives within $K$. Alternatively, $\alpha$ for $T$ can be viewed as weighting a tradeoff between the precision and the absolute number $n_p$ of the selected relevant documents. If $n_p$ is replaced by $\frac{n_p}{N_p}$ in Equation~\ref{eq:T}, then $T$ becomes an arithmetic weighted average of the precision and recall, similar to $F$ being a harmonic weighted average of the precision and recall. Thus, the motivation for $T$, as it is defined by Equation~\ref{eq:T}, is simple removal of unknown $N_p$.

An unnormalized version of $T$
\begin{equation} \label{eq:Tu}
T_u = (1-\alpha)n_p - \alpha n_n
\end{equation}
is simpler and equivalent to $T$ (up to reparameterization by $\alpha$), but we do not have a motivation to remove the normalization of $n_n$ by $K$. If $K$ is changing (for example, in an empirical setting, depending on amount of retrieved texts or their vicinity to the query), the normalization should be helpful. Indeed, in our observations we will see that $T_u$ is not as good a measure as $T$.

\begin{figure*}[t!]
    \centering
    \includegraphics[width=\textwidth]{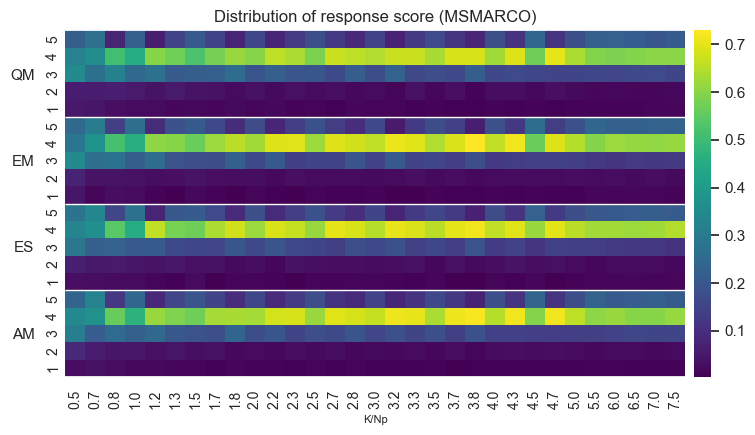}
    \caption{Distribution of the response score (1 to 5) for embedding models shown on Y-axis. On MSMARCO; using only segments with minimum 300 samples, the ratio $\frac{K}{N_p}$ is rounded to the first digit.}
    \label{fig:Grade_DatM}
\end{figure*}

\begin{figure*}[t!]
    \centering
    \includegraphics[width=\textwidth]{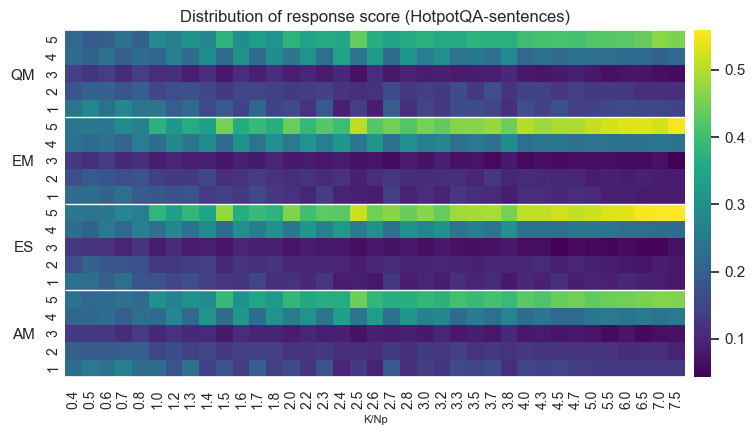}
    \caption{Distribution of the response score (1 to 5) for embedding models shown on Y-axis. On HotpotQA-sentences; using only segments with minimum 300 samples, the ratio $\frac{K}{N_p}$ is rounded to the first digit.}
    \label{fig:Grade_DatHs}
\end{figure*}

\subsection{Embeddings}\label{ssec:Embeddings}
We use the following embedding models (the short notations are for our plots and for our dataset):
\begin{enumerate}[topsep=0pt,itemsep=-1ex,partopsep=1ex,parsep=1ex]
    \item AM\,(AML12v2): all-MiniLM-L12-v2\footnote{https://huggingface.co/sentence-transformers/all-MiniLM-L12-v2} \cite{reimers-gurevych-2019-sentence}
    \item EM\,(ME5S): multilingual-e5-small\footnote{https://huggingface.co/intfloat/multilingual-e5-small} \cite{wang2024multilingual}
    \item ES\,(E5Sv2): e5-small-v2\footnote{https://huggingface.co/intfloat/e5-small-v2} \cite{wang2022text}
    \item QM\,(MQML6Cv1): multi-qa-MiniLM-L6-cos-v1\footnote{https://huggingface.co/sentence-transformers/multi-qa-MiniLM-L6-cos-v1}
\end{enumerate}
All these embeddings have size 384.

\begin{figure*}[t!]
    \centering
    \includegraphics[width=\textwidth]{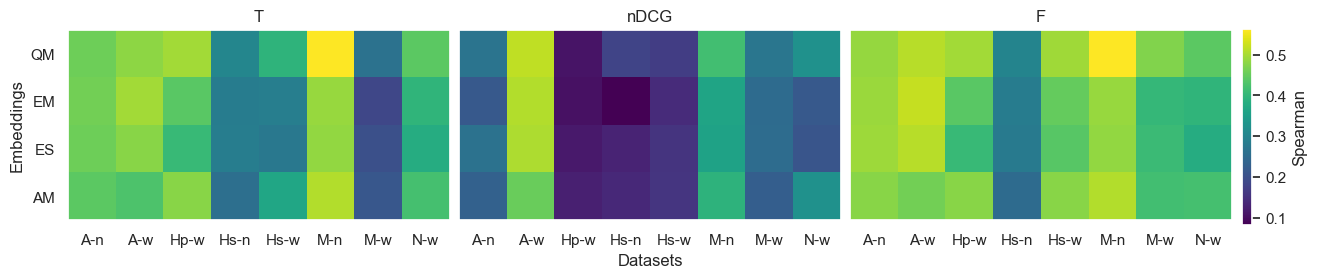} % [width=0.5\textwidth]
    \caption{Spearman correlation between the retrieval measures ($T$, $nDCG$ and $F$) and the response score.}
    \label{fig:SEGcs_comp3_mQ_mNDCG_mF_corr_s}
\end{figure*}

\begin{figure*}[t!]
    \centering
    \includegraphics[width=\textwidth]{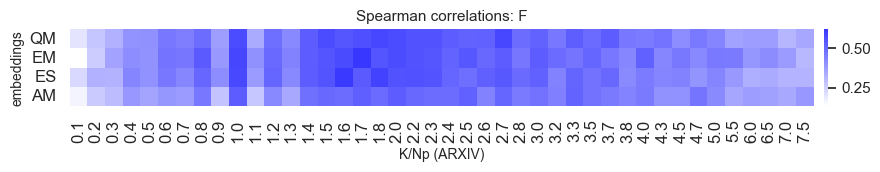} % [width=0.5\textwidth]
    \caption{Spearman correlation between $F$ and the response score, on ARXIV.}
    \label{fig:SEGKtoNp_corr_mF_corr_s_DatA}
\end{figure*}

\subsection{Datasets}\label{ssec:Datasets}
%We sample several well-known datasets to create a single, combine benchmark which we refer to as retrieval-response\footnote{https://huggingface.co/datasets/primer-ai/retrieval-response} (the short notations are for some of our plots):
We sample several well-known datasets to create a benchmark which we refer to as retrieval-response\footnote{https://huggingface.co/datasets/primer-ai/retrieval-response} (the short notations are for some of our plots):
%We sample several well-known datasets to create a single, combine benchmark which we refer to as retrieval-response\footnote{https://zenodo.org/records/18136333?preview=1\&token=eyJhbGciOiJIUzUxMiJ9.eyJpZCI6IjJkMGRlYzY1LTVlOWMtNGQ1ZC1hOThlLWRkNTZhOGRiMDAwOSIsImRhdGEiOnt9LCJyYW5kb20iOiI0MDUxMDAxMDQ2NTI1MGQwMGIyNWQ0YWRlOWM3NGE1YyJ9.jbTu7i0YfEa-aAdsHoZ1LM_nVDwPN2UFDBz4uXpe7QdUDasLcjDQpog0bX9bqkt301-a1fOo1jxeeosihMR6lQ} (the short notations are for some of our plots):

\begin{enumerate}[topsep=0pt,itemsep=-1ex,partopsep=1ex,parsep=1ex]
    \item A: ARXIV (snapshot 173)\footnote{https://huggingface.co/datasets/arxiv-community/arxiv\_dataset} \footnote{https://www.kaggle.com/datasets/Cornell-University/arxiv}
    \item Hp: HotpotQA paragraphs
    \item Hs: HotpotQA sentences
    \item M: MSMARCO \cite{nguen2018msmarco}
    \item N: Natural Questions \cite{kwiatkowski-etal-2019-natural}
\end{enumerate}
 The HotpotQA dataset \cite{yang-etal-2018-hotpotqa} is used with two different granularities: the positives and negatives (by the relevance to a query) are either (1) paragraphs (samples ``Hp'') or (2) sentences (samples ``Hs'').
The full details of the combined dataset are in Appendix~\ref{sapp:data}.
 
In practice, it is generally not known which choice of $K$ will be optimal for a given retrieval. For different queries, it may happen that $K=n_p+n_n$ is either less or greater than the total number $N_p$ of positives. In what follows, we attempt to vary $K$ between values above and below $N_p$ when we have sufficient data.

\subsection{Quality judgment}\label{ssec:judgment}
We assess the quality of each measure by how well it correlates with an LLM-scored quality of a response generated by LLM from a set of top $K$ documents. For each entry in our dataset, the response is compared to an ``ideal'' one, which is based on all the positive documents (see Figure~\ref{fig:schema_all}). This comparison forms the basis of the quality score (Linkert scale 1 to 5, see Appendices~\ref{sapp:prompts_generation} and \ref{sapp:prompts_score}). We use GPT-4o-mini for both generating and scoring the responses.
A higher correlation with the score provides better support for using one measure over another.  For scores which use $\alpha$ (such as $F$ and $T$), the correlations are given by the maximal values as $\alpha$ is varied within a subset of the data (for details see Appendix~\ref{sapp:corr_max_alpha}.

\subsection{LLM vs Human quality scores}\label{ssec:human_scoring}
The judgment by LLM correlates well with human judgment. For 96 random samples, scored by each author using the same prompt and scale as LLM (Appendix~\ref{sapp:prompts_score}), the Krippendorff alpha (ordinal) coefficient is good both between the human scores and between LLM and human scores, as shown in Table~\ref{tab:human_llm_krippendorff_ordinal}.

\begin{table}[th!]
%\small
\centering
\begin{tabular}{@{}c|rrrr@{}}
\hline
{}&{LLM}&{H1}&{H2}&{H3}\\
\hline
	LLM&&0.77&0.77&0.82\\
	H1&0.77&&0.90&0.77\\
	H2&0.77&0.90&&0.77\\
	H3&0.82&0.77&0.77&\\
\hline
\end{tabular}
\caption{Krippendorff alpha (ordinal) coefficient for the LLM score and three human scores (H1, H2, H3), taken on 96 samples.}
\label{tab:human_llm_krippendorff_ordinal}
\end{table}

For comparison, Spearman correlations on the same samples are given in Appendix~\ref{sapp:human_scoring}.

\begin{table*}[t]
\centering
\renewcommand{\arraystretch}{1.8}
\begin{tabular}{ c|c|l|l|l|l|l|l|l|l }
\hline
\hline
\multicolumn{2}{ c| }{$K/N_p$} & \multicolumn{1}{ c| }{0.0 - 1.0} & \multicolumn{1}{ c| }{1.0 - 2.0} & \multicolumn{1}{ c| }{2.0 - 3.0} & \multicolumn{1}{ c| }{3.0 - 4.0} & \multicolumn{1}{ c| }{4.0 - 5.0} & \multicolumn{1}{ c| }{5.0 - 6.0} & \multicolumn{1}{ c| }{6.0 - 7.0} & \multicolumn{1}{ c }{7.0 - } \\
\hline

% ARXIV SECTION
\multirow{4}{*}{\rotatebox[origin=c]{90}{ARXIV}}
& QM & \ccell{mQ}{0.47}{0.38} & \ccell{mQ}{0.72}{0.51} & \ccell{mQ}{0.78}{0.54} & \ccell{mNDCG}{0.78}{0.54} & \ccell{mNDCG}{0.72}{0.51} & \ccell{mNDCG}{0.68}{0.49} & \ccell{mNDCG}{0.70}{0.50} & \ccell{mNDCG}{0.69}{0.49} \\
& EM & \ccell{mQ}{0.48}{0.39} & \ccell{mQ}{0.75}{0.53} & \ccell{mNDCG}{0.78}{0.54} & \ccell{mNDCG}{0.72}{0.51} & \ccell{mNDCG}{0.71}{0.51} & \ccell{mNDCG}{0.70}{0.50} & \ccell{mNDCG}{0.64}{0.47} & \ccell{mNDCG}{0.60}{0.45} \\
& ES & \ccell{mQ}{0.49}{0.39} & \ccell{mF}{0.74}{0.52} & \ccell{mQ}{0.75}{0.53} & \ccell{mNDCG}{0.72}{0.51} & \ccell{mNDCG}{0.68}{0.49} & \ccell{mNDCG}{0.72}{0.51} & \ccell{mNDCG}{0.65}{0.48} & \ccell{mNDCG}{0.68}{0.49} \\
& AM & \ccell{mQ}{0.37}{0.34} & \ccell{mF}{0.60}{0.45} & \ccell{mQ}{0.68}{0.49} & \ccell{mNDCG}{0.66}{0.48} & \ccell{mNDCG}{0.67}{0.48} & \ccell{mNDCG}{0.59}{0.44} & \ccell{mNDCG}{0.61}{0.46} & \ccell{mNDCG}{0.54}{0.42} \\
\hline

% HotPotQA SECTION
\multirow{4}{*}{\rotatebox[origin=c]{90}{HotPotQA}}
& QM & \ccell{mQ}{0.30}{0.30} & \ccell{mQ}{0.63}{0.47} & \ccell{mQ}{0.57}{0.44} & \ccell{mF}{0.62}{0.46} & \ccell{mF}{0.67}{0.48} & \ccell{mF}{0.71}{0.51} & \ccell{mF}{0.76}{0.53} & \ccell{mF}{0.75}{0.52} \\
& EM & \ccell{mQ}{0.28}{0.29} & \ccell{mQ}{0.56}{0.43} & \ccell{mF}{0.47}{0.39} & \ccell{mF}{0.54}{0.42} & \ccell{mF}{0.53}{0.41} & \ccell{mF}{0.58}{0.44} & \ccell{mF}{0.53}{0.42} & \ccell{mF}{0.43}{0.37} \\
& ES & \ccell{mQ}{0.27}{0.29} & \ccell{mQ}{0.54}{0.42} & \ccell{mF}{0.47}{0.38} & \ccell{mF}{0.51}{0.41} & \ccell{mF}{0.47}{0.38} & \ccell{mF}{0.48}{0.39} & \ccell{mF}{0.48}{0.39} & \ccell{mF}{0.33}{0.31} \\
& AM & \ccell{mQ}{0.21}{0.25} & \ccell{mQ}{0.57}{0.43} & \ccell{mQ}{0.52}{0.41} & \ccell{mF}{0.59}{0.45} & \ccell{mF}{0.66}{0.48} & \ccell{mF}{0.71}{0.50} & \ccell{mF}{0.75}{0.53} & \ccell{mF}{0.65}{0.48} \\
\hline

% MSMARCO SECTION
\multirow{4}{*}{\rotatebox[origin=c]{90}{MSMARCO}}
& QM & \ccell{mF}{0.85}{0.57} & \ccell{mF}{0.90}{0.60} & \ccell{mF}{0.71}{0.50} & \ccell{mF}{0.54}{0.42} & \ccell{mF}{0.46}{0.38} & \ccell{mF}{0.33}{0.31} & \ccell{mF}{0.29}{0.30} & \ccell{mF}{0.30}{0.30} \\
& EM & \ccell{mF}{0.73}{0.51} & \ccell{mF}{0.74}{0.52} & \ccell{mF}{0.49}{0.40} & \ccell{mF}{0.45}{0.38} & \ccell{mF}{0.30}{0.30} & \ccell{mF}{0.27}{0.29} & \ccell{mF}{0.28}{0.29} & \ccell{mF}{0.21}{0.26} \\
& ES & \ccell{mF}{0.72}{0.51} & \ccell{mF}{0.81}{0.56} & \ccell{mF}{0.56}{0.43} & \ccell{mF}{0.41}{0.36} & \ccell{mF}{0.36}{0.33} & \ccell{mF}{0.24}{0.27} & \ccell{mF}{0.17}{0.23} & \ccell{mF}{0.22}{0.26} \\
& AM & \ccell{mF}{0.78}{0.54} & \ccell{mF}{0.81}{0.56} & \ccell{mF}{0.57}{0.43} & \ccell{mF}{0.41}{0.36} & \ccell{mF}{0.28}{0.29} & \ccell{mF}{0.27}{0.29} & \ccell{mF}{0.22}{0.26} & \ccell{mF}{0.23}{0.27} \\
\hline

% Natural Questions SECTION
\multirow{4}{*}{\rotatebox[origin=c]{90}{Natural Questions}}
& QM & \textbf{-} & \ccell{mF}{0.89}{0.59} & \ccell{mF}{0.74}{0.52} & \ccell{mF}{0.57}{0.44} & \ccell{mF}{0.42}{0.36} & \ccell{mF}{0.36}{0.33} & \ccell{mF}{0.34}{0.32} & \ccell{mF}{0.30}{0.30} \\
& EM & \textbf{-} & \ccell{mF}{0.80}{0.55} & \ccell{mF}{0.58}{0.44} & \ccell{mF}{0.46}{0.38} & \ccell{mF}{0.40}{0.35} & \ccell{mF}{0.30}{0.30} & \ccell{mF}{0.23}{0.27} & \ccell{mF}{0.20}{0.25} \\
& ES & \textbf{-} & \ccell{mF}{0.73}{0.52} & \ccell{mF}{0.55}{0.43} & \ccell{mF}{0.47}{0.38} & \ccell{mF}{0.37}{0.33} & \ccell{mF}{0.23}{0.26} & \ccell{mF}{0.14}{0.22} & \ccell{mF}{0.18}{0.24} \\
& AM & \textbf{-} & \ccell{mF}{0.81}{0.55} & \ccell{mF}{0.63}{0.47} & \ccell{mF}{0.58}{0.44} & \ccell{mF}{0.44}{0.37} & \ccell{mF}{0.40}{0.35} & \ccell{mF}{0.31}{0.30} & \ccell{mF}{0.31}{0.30} \\
\hline
\hline
\end{tabular}

\caption{Maximum spearman correlations of all the measures.  Colors represent the measure which achieves the highest correlation for a given dataset, embedding model and range of $K / N_p$.  {\color{mQ}$\blacksquare$} represents $T$.  {\color{mNDCG}$\blacksquare$} represents $nDCG$.  {\color{mF}$\blacksquare$} represents $F$.
}
\label{tab:sum}
\end{table*}

\begin{figure*}[t!]
    \centering
    \includegraphics[width=\textwidth]{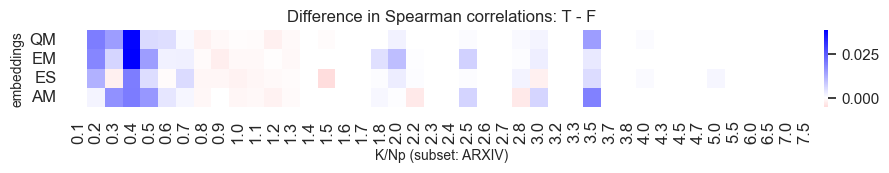} % [width=0.5\textwidth]
    \caption{Difference between the Spearman correlations: $T$-response minus $F$-response. On ARXIV; using only segments with minimum 300 samples, the ratio $\frac{K}{N_p}$ is rounded to the first digit.}
    \label{fig:SEGKtoNp_diff_mF_mQ_corr_s_DatA}
\end{figure*}
\begin{figure*}[t!]
    \centering
    \includegraphics[width=\textwidth]{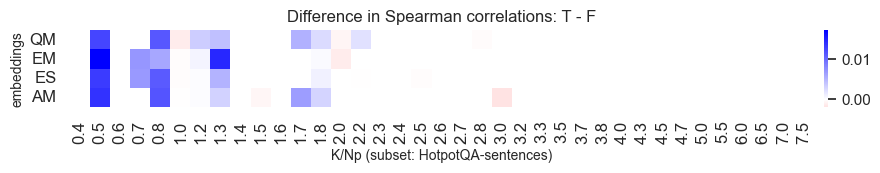} % [width=0.5\textwidth]
    \caption{Difference between the Spearman correlations: $T$-response minus $F$-response. On HotpotQA-sentences.}
    \label{fig:SEGKtoNp_diff_mF_mQ_corr_s_DatHs}
\end{figure*}
\begin{figure*}[t!]
    \centering
    \includegraphics[width=\textwidth]{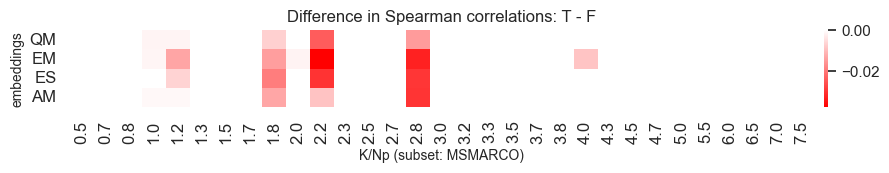} % [width=0.5\textwidth]
    \caption{Difference between the Spearman correlations: $T$-response minus $F$-response. On MSMARCO.}
    \label{fig:SEGKtoNp_diff_mF_mQ_corr_s_DatM}
\end{figure*}
\begin{figure*}[t!]
    \centering
    \includegraphics[width=\textwidth]{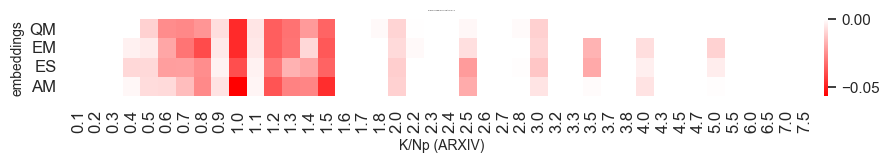} % [width=0.5\textwidth]
    \caption{Difference between the Spearman correlations: $T_u$-response minus $T$-response. On ARXIV.}
    \label{fig:SEGKtoNp_diff_mQ_mQu_corr_s_DatA}
\end{figure*}
\begin{figure*}[t!]
    \centering
    \includegraphics[width=\textwidth]{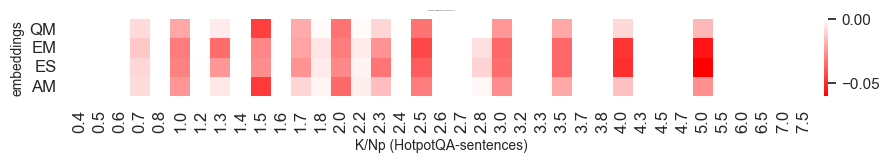} % [width=0.5\textwidth]
    \caption{Difference between the Spearman correlations: $T_u$-response minus $T$-response. On HotpotQA-sentences.}
    \label{fig:SEGKtoNp_diff_mQ_mQu_corr_s_DatHs}
\end{figure*}
\begin{figure*}[t!]
    \centering
    \includegraphics[width=\textwidth]{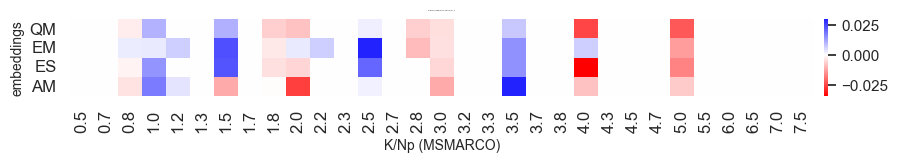} % [width=0.5\textwidth]
    \caption{Difference between the Spearman correlations: $T_u$-response minus $T$-response. On MSMARCO.}
    \label{fig:SEGKtoNp_diff_mQ_mQu_corr_s_DatM}
\end{figure*}

\section{Observations}\label{sec:Observations}
\subsection{Response score distribution}\label{ssec:response_score}
The distribution of the response score is shown for ARXIV in Figure~\ref{fig:Grade_DatA}. We split the dataset into subsets according to the ratio $K$/$N_p$, rounded to the first digit, and retain only the subsets which have at least 300 samples. In each subset (and for each embedding model) the sum of the values (5 vertical cells in the heatmap) equals 1.

We see that for ARXIV, the best scored responses occur at values of $K$ which are comparable to $N_p$. At lower values of $K/N_p$, there may not be enough positive texts to give a good response. At high $K/N_p$ there may be too many negative texts within $K$, which can confuse the LLM and lead to lower scores. For ARXIV specifically, these scores may be exacerbated due to the difficulty of the dataset.

Figure~\ref{fig:Grade_DatM} shows a similar pattern for MSMARCO, but with a higher ``optimal'' ratio of $\frac{K}{N_p}$ than for ARXIV. We speculate the reason for this is that MSMARCO contains texts which are simpler to understand relative to ARXIV.  Therefore the detriment which comes from adding extra negatives is outweighed by the increasing likelihood of capturing more positives. This is demonstrated to a further extent with HotpotQS-sentences (Figure~\ref{fig:Grade_DatHs}, which must be even simpler for the LLM to understand given the higher ``optimal'' value of $K/N_P$.

Our observations for the Natural Questions and HotpotQA-paragraphs datasets are similar to HotpotQA-sentences, but have fewer available segments; they are shown in Appendix~\ref{app:response_score}.

\begin{figure*}[t!]
    \centering
    \includegraphics[width=\textwidth]{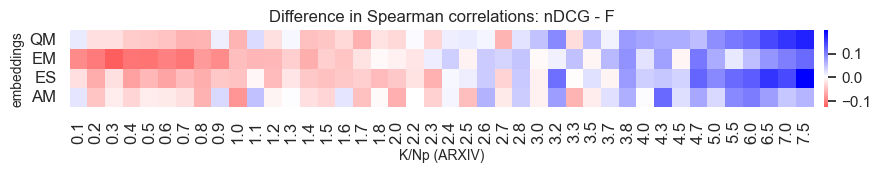} % [width=0.5\textwidth]
    \caption{Difference between the Spearman correlations: $nDCG$-response minus $F$-response. On ARXIV.}
    \label{fig:SEGKtoNp_diff_mF_mNDCG_corr_s_DatA}
\end{figure*}
\begin{figure*}[t!]
    \centering
    \includegraphics[width=\textwidth]{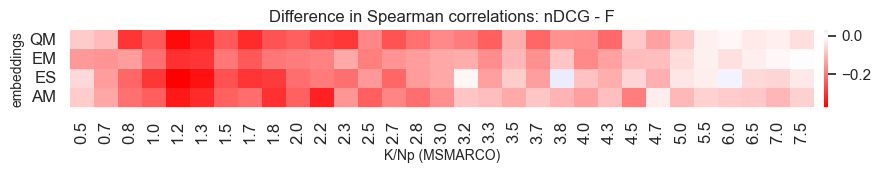} % [width=0.5\textwidth]
    \caption{Difference between the Spearman correlations: $nDCG$-response minus $F$-response. On MSMARCO.}
    \label{fig:SEGKtoNp_diff_mF_mNDCG_corr_s_DatM}
\end{figure*}
\begin{figure*}[t!]
    \centering
    \includegraphics[width=\textwidth]{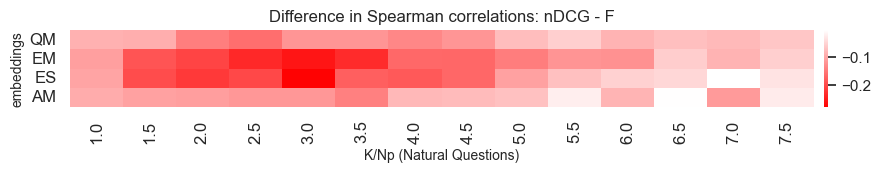} % [width=0.5\textwidth]
    \caption{Difference between the Spearman correlations: $nDCG$-response minus $F$-response. On Natural Questions.}
    \label{fig:SEGKtoNp_diff_mF_mNDCG_corr_s_DatN}
\end{figure*}
\begin{figure}[t!]
    \centering
    \includegraphics[width=0.5\textwidth]{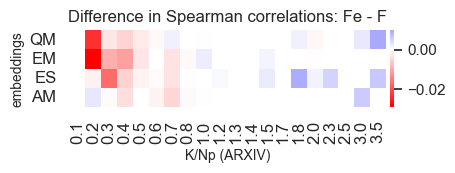} % [width=0.5\textwidth]
    \caption{Difference between the Spearman correlations: $F_e$-response minus $F$-response. On ARXIV.}
    \label{fig:SEGKtoNp_diff_mF_mE_corr_s_DatA}
\end{figure}
\begin{figure}[t!]
    \centering
    \includegraphics[width=0.5\textwidth]{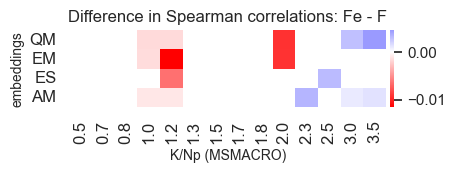} % [width=0.5\textwidth]
    \caption{Difference between the Spearman correlations: $F_e$-response minus $F$-response. On MSMARCO.}
    \label{fig:SEGKtoNp_diff_mF_mE_corr_s_DatM}
\end{figure}
\begin{figure*}[t!]
    \centering
    \includegraphics[width=\textwidth]{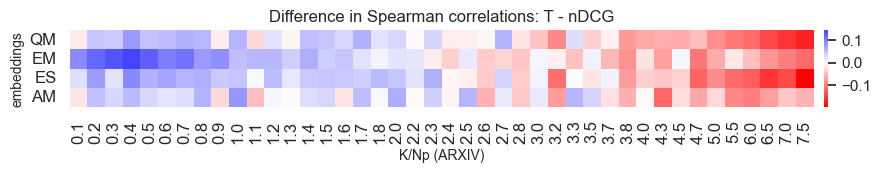} % [width=0.5\textwidth]
    \caption{Difference between the Spearman correlations: $Q$-response minus $nDCG$-response. On ARXIV.}
    \label{fig:SEGKtoNp_diff_mNDCG_mQ_corr_s_DatA}
\end{figure*}

\subsection{Correlations}\label{ssec:correlations}
Spearman correlations of $F$, $nDCG$ and $T$ with the response score are shown in Figure~\ref{fig:SEGcs_comp3_mQ_mNDCG_mF_corr_s}.
The datasets defined in Section~\ref{ssec:Datasets} are split here into ``narrow'' (suffix ``-n'') and ``wide'' (suffix ``-w''), with samples having $K$<$N_p$ and $K \geq N_p$ respectively. The Pearson and Kendall Tau correlations have similar trends and are given in Appendix~\ref{app:Correlations}.

From Figure~\ref{fig:schema_all} we can see that for many of the segments considered, the $T$ measure does not lose much in terms of correlations compared to $F$ even though it is formulated without $N_p$. 

For a more detailed analysis, we consider more refined segments of data - splitting the datasets by the ratio $\frac{K}{N_p}$, rounded to the first digit, as described in Section~\ref{ssec:response_score}. Examples of the correlations from ARXIV on such segments are shown in Figure~\ref{fig:SEGKtoNp_corr_mF_corr_s_DatA} for the $F$ measure.
The level of correlations between $F$ and the response score is similar for other datasets (Appendix~\ref{sapp:corr_KtoNp}). The correlations between $T$ and the response score are also not very different (Appendix~\ref{sapp:corr_KtoNp}). For this reason, in the next Section we show the differences between the correlations.

\subsection{Differences between measures}\label{ssec:differences}

Table~\ref{tab:sum} gives an overview of the Spearman correlations for $T$, $F$ and $nDCG$ as the dataset, embedding model and $K / N_p$ are varied.  In general, we see that $T$ performs best for ARXIV and HotPotQA-sentences in the low $K / N_p$ regime and is overtaken by $nDCG$ and $F$ as $K / N_p$ increases.  For MSMARCO and Natural Questions, $F$ appears to outperform both $T$ and $nDCG$ for all values of $K / N_p$ tested.

Figures~\ref{fig:SEGKtoNp_diff_mF_mQ_corr_s_DatA},~\ref{fig:SEGKtoNp_diff_mF_mQ_corr_s_DatHs} and~\ref{fig:SEGKtoNp_diff_mF_mQ_corr_s_DatM} show the difference between the correlations of $T$ with the response score and $F$ with the response score. The difference depends more on the data than on the embeddings.  Specifically, for ARXIV and HotpotQA-sentences, the $T$ measure correlates better with the response score than $F$ on most segments; for MSMARCO it is worse. The results for HotpotQA-paragraphs and Natural Questions are not shown because there is no noticeable difference for them.

The simpler, unnormalized version of $T$ ($T_u$, Equation~\ref{eq:Tu}) has worse correlations with the response quality score on ARXIV (Figure~\ref{fig:SEGKtoNp_diff_mQ_mQu_corr_s_DatA}) and on HotpotQA-sentences (Figure~\ref{fig:SEGKtoNp_diff_mQ_mQu_corr_s_DatHs}), has mixed and less difference on MSMARCO (Figure~\ref{fig:SEGKtoNp_diff_mQ_mQu_corr_s_DatM}) and has no noticeable difference on HotpotQA-paragraphs and on Natural Questions.

For the majority of the datasets, $nDCG$ shows worse correlations relative to $F$, but gets better for high $\frac{K}{N_p}$ in ARXIV (Figure~\ref{fig:SEGKtoNp_diff_mF_mNDCG_corr_s_DatA}).

We suggest the reason lies in the complexity of the ARXIV abstracts, which yields greater benefits when the top $K$ findings are presented to the LLM in an easier order: first relevant, then non-relevant. The order becomes especially important for high $\frac{K}{N_p}$ ratio, when there are inevitably many negative (irrelevant) items within top $K$. A similar trend can be seen for MSMARCO (Figure~\ref{fig:SEGKtoNp_diff_mF_mNDCG_corr_s_DatM}) and Natural Questions (Figure~\ref{fig:SEGKtoNp_diff_mF_mNDCG_corr_s_DatN}); however for these datasets, $nDCG$ is still worse than $F$. The trend is less pronounced for HotpotQA (see Figures~\ref{fig:SEGKtoNp_diff_mF_mNDCG_corr_s_DatHs} and~\ref{fig:SEGKtoNp_diff_mF_mNDCG_corr_s_DatHp} and Appendix~\ref{app:Differences}), possibly because HotpotQA-sentences samples are too simple and because the range of $K/N_p$ for HotpotQA-paragraphs is too narrow.

In Figures~\ref{fig:SEGKtoNp_diff_mF_mE_corr_s_DatA} and \ref{fig:SEGKtoNp_diff_mF_mE_corr_s_DatM}, we see similar trend appearing for $F_e$ ($F$ with $N_p$ replaced by a crude estimate - $n_p$ within $2K$) when evaluating on ARXIV and MSMARCO.
The reason here, we assume, is that the estimate of $N_p$, however primitive, is still more reliable at higher ratios $\frac{K}{N_p}$, simply because the top $2K$ documents have more positives.
The comparisons on other datasets are more limited, and shown in Figures~\ref{fig:EGKtoNp_diff_mF_mE_corr_s_DatHs} and \ref{fig:EGKtoNp_diff_mF_mE_corr_s_DatN} in Appendix~\ref{app:Differences} (no noticeable difference on HotpotQA-paragraphs, not shown).

It is surprising, however, that $F_e$ at high ratios of $\frac{K}{N_p}$ can become even better than $F$, meaning that an estimated $N_p$ serves better than the real $N_p$. We suspect the reason is that the relevant documents falling into the top $2K$ (used in our estimation of $N_p$) are more important and should ``weigh more'' than other relevant documents when counting $N_p$. Because these documents happen to be closer to the query (by embedding similarity), it is likely they also would be easier for the LLM and more useful, especially at high $\frac{K}{N_p}$.

As we already observed (Figures~\ref{fig:SEGKtoNp_diff_mF_mNDCG_corr_s_DatA}, \ref{fig:SEGKtoNp_diff_mF_mNDCG_corr_s_DatM}, \ref{fig:SEGKtoNp_diff_mF_mNDCG_corr_s_DatN}, \ref{fig:SEGKtoNp_diff_mF_mNDCG_corr_s_DatHs}, \ref{fig:SEGKtoNp_diff_mF_mNDCG_corr_s_DatHp}), $nDCG$ is mostly worse that $F$, except at high $\frac{K}{N_p}$ values for ARXIV - when the content is difficult and the order of numerous negatives (irrelevant documents) within top $K$ is important for the LLM. Since $T$ is better or not too different from $F$ (Figures~\ref{fig:SEGKtoNp_diff_mF_mQ_corr_s_DatA}, ~\ref{fig:SEGKtoNp_diff_mF_mQ_corr_s_DatHs} and~\ref{fig:SEGKtoNp_diff_mF_mQ_corr_s_DatM}), $T$ is also better than $nDCG$ at not too high $\frac{K}{N_p}$ values for ARXIV (Figure~\ref{fig:SEGKtoNp_diff_mNDCG_mQ_corr_s_DatA}); it is also better than $nDCG$ for other datasets (Figures~\ref{fig:SEGKtoNp_diff_mNDCG_mQ_corr_s_DatHs}, \ref{fig:SEGKtoNp_diff_mNDCG_mQ_corr_s_DatHp}, \ref{fig:SEGKtoNp_diff_mNDCG_mQ_corr_s_DatM} and \ref{fig:SEGKtoNp_diff_mNDCG_mQ_corr_s_DatN} in Appendix~\ref{app:Differences}).

\section{Conclusion}\label{sec:Conclusion}
We considered a few measures representing different approaches to using $N_p$, the total count of documents relevant to a query. Measure $F$ requires knowing $N_p$, which can be achieved by labeling and fixing part of KB. Measures $T$ (suggested here) and $nDCG$ do not use $N_p$. Measure $F_e$ uses a crudely estimated value for $N_p$. We compared how well these measures correlate with a quality score, assigned (by an LLM) to a response generated (by an LLM) from the top $K$ retrieved documents. Our conclusions:
\begin{enumerate}[topsep=0pt,itemsep=-1ex,partopsep=1ex,parsep=1ex]
    \item Only at high $\frac{K}{N_p}$ ratios, as the order of documents within top-$K$ becomes too important for the LLM, $nDCG$ may become better than $F$ or $T$ on a dataset with complicated enough documents (ARXIV).
    %\item At $\frac{K}{N_p}$ ratios comparable to $1$ or lower, the $T$ measure is better than $nDCG$. For ratios below $1$, it is also better or the same as $F$, depending on the type of the documents.
    \item Except the above, the $T$ measure is better than $nDCG$, and similar to $F$ (and at low or moderate $\frac{K}{N_p}$ can be better, depending on a dataset).
    \item At high $\frac{K}{N_p}$ ratios, substituting a simple estimate of $N_p$ from a $2K$ selection instead of the real $N_p$ not just approximates $F$, but even improves it. We suggest the reason is that the relevant documents ranked higher (and therefore falling into the top $2K$) would be more important for the LLM response than the rest of the relevant documents.
    \item Within our observations, the document type and the ratio $\frac{K}{N_p}$ appear to be more important than the choice of an embedding model when selecting a retrieval measure (and simply for the response score).
\end{enumerate}
Of course in practice, the range of $\frac{K}{N_p}$ depends on the distribution of $N_p$ in the used data, the choice of $K$ (which may be varied at runtime depending on a query and the retrieved texts) and on a consideration of computation expenses and time lag (growing with $K$).

%\clearpage
\section*{Limitations}\label{sec:Limitations}

Our experiments focus on retrieval scenarios with relatively moderate ranges of $N_p$, which may limit the applicability of our findings to broader retrieval regimes.

We left out a consideration of the additional latency incurred when evaluating the results of a retrieval in realtime. The measures (even not involving a knowledge of $N_p$) require identifying relevancy of the top-$K$ documents with respect to a query. Outside of using a fixed query with a corresponding set of labeled documents, one direct method of obtaining labels is by using an LLM to assess the relevancy of each document to the query after making the top-$K$ selection.  This is particularly useful in a retrieval system monitoring setting, where any additional time spent measuring the quality of a retrieval will not block any downstream tasks--the system simply continues on while the results are recorded in the background.

%\bibliography{anthology,custom}
%\bibliographystyle{acl_natbib}
\bibliography{custom}

\appendix

\begin{figure}[t!]
    \centering
    \includegraphics[width=0.5\textwidth]{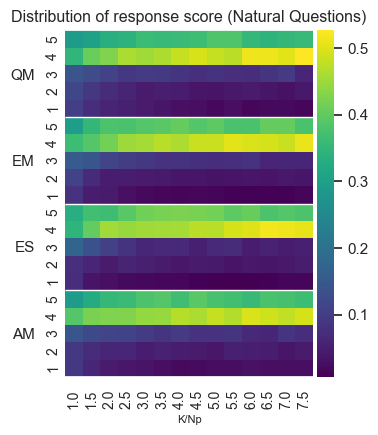}
    \caption{Distribution of the response score (1 to 5) for embedding models shown on Y-axis. On Natural Questions; using only segments with minimum 300 samples, the ratio $\frac{K}{N_p}$ is rounded to the first digit.}
    \label{fig:Grade_DatN}
\end{figure}

\begin{figure}[t!]
    \centering
    \includegraphics[width=0.5\textwidth]{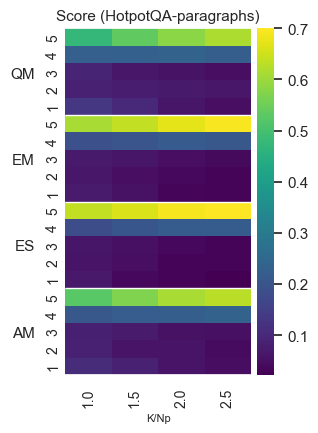}
    \caption{Distribution of the response score (1 to 5) for embedding models shown on Y-axis. On HotpotQA-paragraphs; using only segments with minimum 300 samples, the ratio $\frac{K}{N_p}$ is rounded to the first digit.}
    \label{fig:Grade_DatHp}
\end{figure}

\section{Dataset Retrieval-Response}\label{app:dataset}
\subsection{Data}\label{sapp:data}
% The dataset as described here:
% https://drive.google.com/drive/folders/1aRO7qZKFgN_zwMa_LB88xFXt5g8w8-Bo
%The dataset\footnote{https://huggingface.co/datasets/primer-ai/retrieval-response} consists of three parts:
The dataset\footnote{https://huggingface.co/datasets/primer-ai/retrieval-response} consists of four parts:
\begin{enumerate}[topsep=0pt,itemsep=-1ex,partopsep=1ex,parsep=1ex]
    \item Query-texts samples
    \item Ranked samples
    \item Graded samples
    \item Human-graded samples
\end{enumerate}
\subsubsection{Query-texts samples}\label{sapp:samples_text}
Query-texts samples consist of 6112 samples, each sample is a dictionary with the following items:
\begin{enumerate}[topsep=0pt,itemsep=-1ex,partopsep=1ex,parsep=1ex]
    \item ``id'': A string Id of the sample. The Id consists of a name of a subset, concatenated by ``-'' with Id of the item in the subset. For example, Id=``N-5'' means that it is sample \#5 from the subset Natural Questions. Each sample is uniquely identified by its Id.
    \item ``q'': The query.
    \item ``p'': The list of positives.
    \item ``n'': The list of negatives.
\end{enumerate}
All the subsets in the dataset: ``A'', ``Hp-e'', ``Hp-h'', ``Hp-m'', ``Hs-e'', ``Hs-h'', ``Hs-m'', ``M'', ``N''. The short names, as explained in Section~\ref{sec:Setup}, are ``A'' for ARXIV, ``H'' for HotpotQA, ``M'' for MSMARCO and ``N'' for Natural Questions.

HotpotQA appears with two different granularities: The positives and negatives are (1) paragraphs in ``Hp'' and (2) sentences in ``Hs''. Beyond that, we keep the HotpotQA classification of the queries: easy (suffix ``-e''), medium (``-m'') and hard (``-h'').

In the ARXIV sample, a query is a name of an arxiv category. The positives are abstracts of papers in this category and the negatives are abstracts of papers in a related, but different category. We kept the following additional, not strictly necessary, information, with additional keys:
\begin{enumerate}[topsep=0pt,itemsep=-1ex,partopsep=1ex,parsep=1ex]
    \item ``c1'': The symbolic name of a category for positives. (The name of this category serves as the query and is stored with the key ``q''.)
    \item ``c2'': The symbolic name of a category for negatives.
    \item ``q2'': The name of the category for negatives (not used).
\end{enumerate}
For example, a sample with id=``A-0'' has c1=``math.ca'', c2=``math.pr'', q=``classical analysis and ODEs'' and q2=``probability'' (it also has a list of positives ``p'' and a list of negatives ``n'').

\subsubsection{Ranked samples}\label{sapp:samples_ranked}
Each of the query-texts samples (Appendix~\ref{sapp:samples_text}) can be used as a retrieval example with different number $N_c$ of candidates, number $N_p$ of positives ($N_p$<$N_c$), number $K$ of top retrieved texts, and the retrieval can be done with different embeddings.

The number of ranked samples is 42992; each of the four embeddings (listed in Section~\ref{sec:Setup}) is used in 10748 samples.
Each ranked sample is a dictionary with the following items:
\begin{enumerate}[topsep=0pt,itemsep=-1ex,partopsep=1ex,parsep=1ex]
    \item ``id'': A string Id of the sample, the same as in the query-texts samples.
    \item ``E'': The embedding's short notation, as specified in Section~\ref{sec:Setup}. The embedding used for ranking of all the candidates and selecting top-K candidates.
    \item ``Nc'': Total number of candidates (positives and negatives), taken from the corresponding query-texts sample (with the same ``id'').
    \item ``Np'': Total number of positives, taken as the first $N_p$ positives ``p'' of the corresponding query-texts sample. (Negatives are also taken as first $N_c$-$N_p$ from the negatives ``n''.)
    \item ``K'': A sorted list of all the $K$ (number of retrieved candidates, ``top-K'') used for this sample.
    \item ``P'': A list of precisions calculated for the top-K specified in the list ``K'', in the same order. Has the same length as the list ``K''.
    \item ``R'': A list of recalls calculated for the top-K specified in the list ``K'', in the same order.
    \item ``rank'': A list (length $N_c$) of indexes of all the candidates, sorted by ranks accordingly to cosine similarities with query, by the embedding ``E''.
\end{enumerate} 
Each ranked sample is uniquely identified by the tuple ($id$, $E$, $N_c$, $N_p$).
 
In order to get the ranked texts corresponding to the ``rank'' list of a ranked sample $S_r$, its query-texts sample $S_q$ (the sample with the same ``id'') can be used as in this snippet:
\begin{lstlisting}
Np = Sr['Np']
texts = [
    (Sq['p'][i] if i<Np else Sq['n'][i-Np]) 
    for i in Sr['rank']]
\end{lstlisting}

To assure an understanding of the data of a ranked sample $S_r$ and of the corresponding query-texts sample $S_q$, see the following assertions:
\begin{lstlisting}
len(Sr['P']) == len(Sr['K'])
len(Sr['R']) == len(Sr['K'])
len(Sr['rank']) == Sr['Nc']
Sr['Nc'] <= len(Sq['p']) + len(Sq['n'])
Sr['Np'] <= len(Sq['p'])
Sr['Np'] < Sr['Nc']
Sr['Nc'] >= 2
Sr['Np'] >= 2
\end{lstlisting}

\begin{figure*}[t!]
    \centering
    \includegraphics[width=\textwidth]{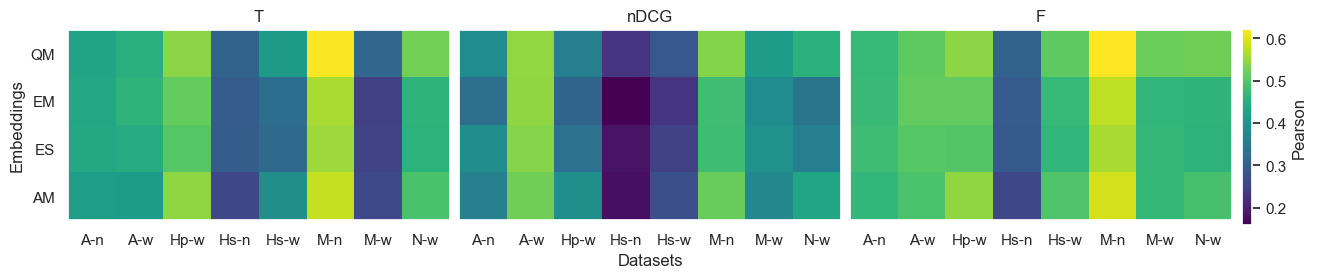} % [width=0.5\textwidth]
    \caption{Pearson correlation between the retrieval measures ($T$, $nDCG$ and $F$) and the response score.}
    \label{fig:SEGcs_comp3_mQ_mNDCG_mF_corr_p}
\end{figure*}
\begin{figure*}[t!]
    \centering
    \includegraphics[width=\textwidth]{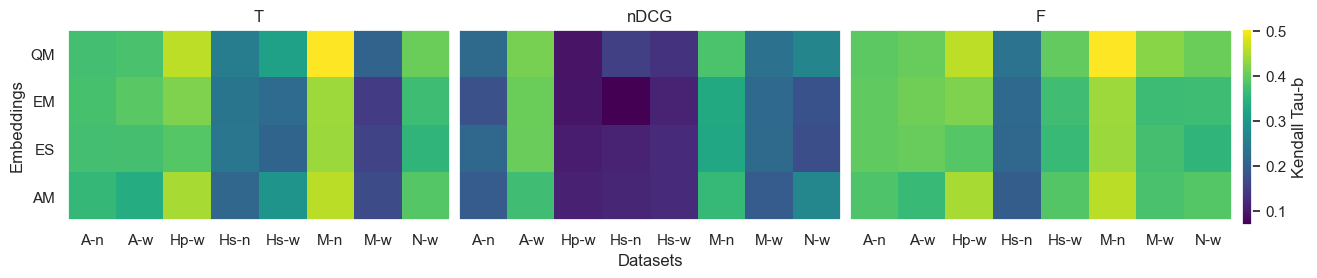} % [width=0.5\textwidth]
    \caption{Kendall Tau-b correlation between the retrieval measures ($T$, $nDCG$ and $F$) and the response score.}
    \label{fig:SEGcs_comp3_mQ_mNDCG_mF_corr_kb}
\end{figure*}
\begin{figure*}[t!]
    \centering
    \includegraphics[width=\textwidth]{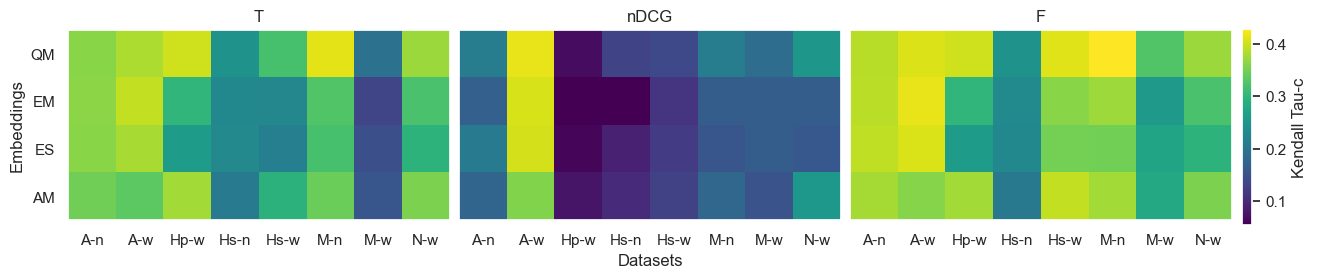} % [width=0.5\textwidth]
    \caption{Kendall Tau-c correlation between the retrieval measures ($T$, $nDCG$ and $F$) and the response score.}
    \label{fig:SEGcs_comp3_mQ_mNDCG_mF_corr_kc}
\end{figure*}

\begin{figure*}[t!]
    \centering
    \includegraphics[width=\textwidth]{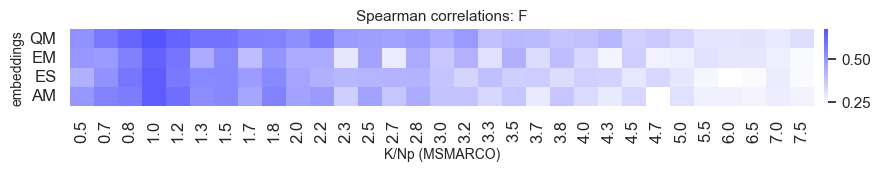}
    \caption{Spearman correlation between $F$ and the response score, on MSMARCO.}
    \label{fig:SEGKtoNp_corr_mF_corr_s_DatM}
\end{figure*}
\begin{figure*}[t!]
    \centering
    \includegraphics[width=\textwidth]{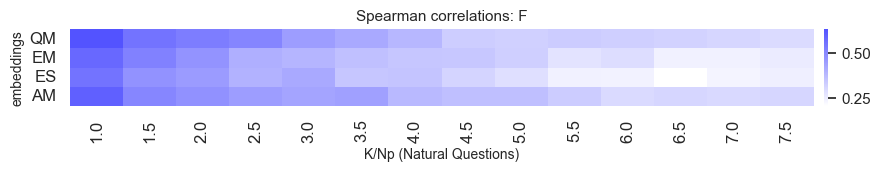}
    \caption{Spearman correlation between $F$ and the response score, on Natural Questions.}
    \label{fig:SEGKtoNp_corr_mF_corr_s_DatN}
\end{figure*}
\begin{figure*}[t!]
    \centering
    \includegraphics[width=\textwidth]{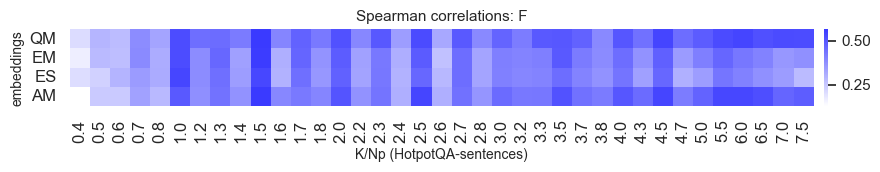}
    \caption{Spearman correlation between $F$ and the response score, on HotpotQA-sentences.}
    \label{fig:SEGKtoNp_corr_mF_corr_s_DatHs}
\end{figure*}
\begin{figure}[t!]
    \centering
    \includegraphics[width=0.5\textwidth]{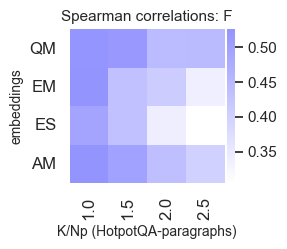}
    \caption{Spearman correlation between $F$ and the response score, on HotpotQA-paragraphs.}
    \label{fig:SEGKtoNp_corr_mF_corr_s_DatHp}
\end{figure}

\begin{figure*}[t!]
    \centering
    \includegraphics[width=\textwidth]{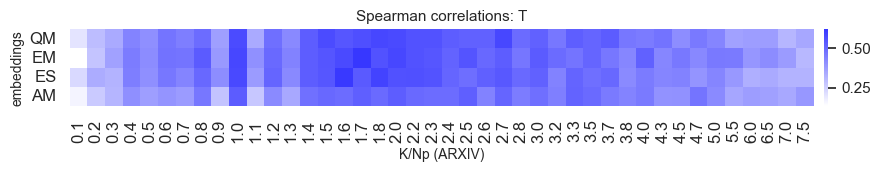}
    \caption{Spearman correlation between $T$ and the response score, on ARXIV.}
    \label{fig:SEGKtoNp_corr_mQ_corr_s_DatA}
\end{figure*}
\begin{figure*}[t!]
    \centering
    \includegraphics[width=\textwidth]{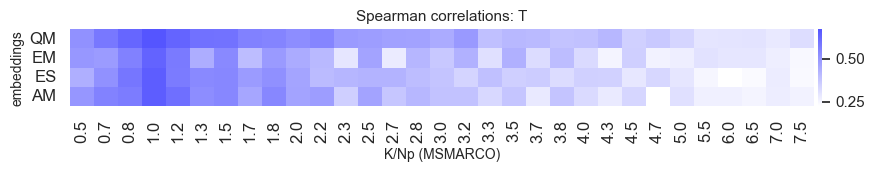}
    \caption{Spearman correlation between $T$ and the response score, on MSMARCO.}
    \label{fig:SEGKtoNp_corr_mQ_corr_s_DatM}
\end{figure*}
\begin{figure*}[t!]
    \centering
    \includegraphics[width=\textwidth]{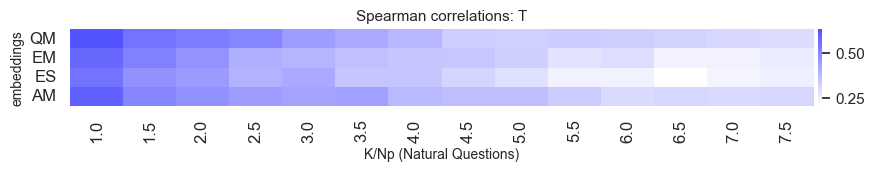}
    \caption{Spearman correlation between $T$ and the response score, on Natural Questions.}
    \label{fig:SEGKtoNp_corr_mQ_corr_s_DatN}
\end{figure*}
\begin{figure*}[t!]
    \centering
    \includegraphics[width=\textwidth]{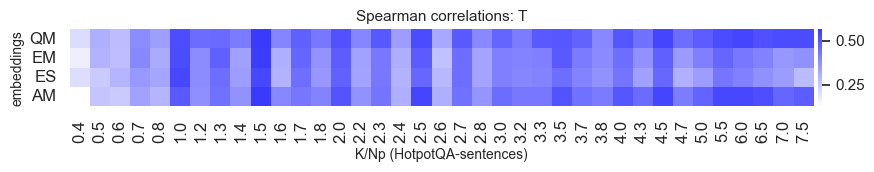}
    \caption{Spearman correlation between $T$ and the response score, on HotpotQA-sentences.}
    \label{fig:SEGKtoNp_corr_mQ_corr_s_DatHs}
\end{figure*}
\begin{figure}[t!]
    \centering
    \includegraphics[width=0.4\textwidth]{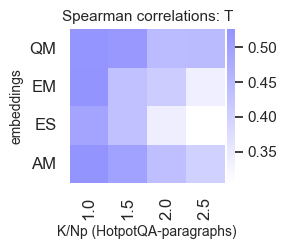}
    \caption{Spearman correlation between $T$ and the response score, on HotpotQA-paragraphs.}
    \label{fig:SEGKtoNp_corr_mQ_corr_s_DatHp}
\end{figure}

\subsubsection{Graded samples}\label{sapp:samples_graded}
The LLM grading is done for each choice of ``top-K'' in each of the ranked samples. This means that each ranked sample $S_r$ makes \textit{len}($S_r$['K']) graded samples. In total, this gives 535888 graded samples.

The grading is done as outlined in Section~\ref{sec:Setup}; more details and the prompts are in Appendices~\ref{sapp:prompts_generation} and \ref{sapp:prompts_score}.
Each graded sample is a dictionary with the following items, most of which are familiar from the ranked samples:
\begin{enumerate}[topsep=0pt,itemsep=-1ex,partopsep=1ex,parsep=1ex]
    \item ``id'': A string Id of the sample, the same as in the query-texts samples.
    \item ``E'': The embedding's short notation.
    \item ``Nc'': Total number of candidates (positives and negatives).
    \item ``Np'': Total number of positives.
    \item ``K'': A value of K (``top-K'') taken from the list of ``K'' in the corresponding ranked sample.
    \item ``rank'': A list equal to the first K elements of the list ``rank'' of the corresponding ranked sample $S_r$, i.e. Sr[``rank''][:K]. 
    \item ``inK'': A list created from the ``rank'' (the item above), by replacement of each index by 1 (if positive) or 0 (if negative). 
    \item ``answer\_ideal'': LLM-generated answer to the query, obtained by using all the positives from the corresponding query-texts sample.
    \item ``answer\_topK'': LLN-generated answer to the query, obtained by using the retrieved $K$ candidates, given to LLM in their ranking order.
    \item ``grade'': The LLM-generated score (on Linkert scale from 1 to 5), obtained by comparing the top-K answer to the ideal answer, with the knowledge of the query.
    \item ``P'': A value of precision corresponding to the selected $K$; given here for convenience.
    \item ``R'': A value of recall corresponding to the selected $K$; given here for convenience.
\end{enumerate}
Each graded sample is uniquely identified by the tuple ($id$, $E$, $N_c$, $N_p$, $K$) and it is related to its ranked sample by the tuple ($id$, $E$, $N_c$, $N_p$).
To assure an understanding of the data of a graded sample $S_g$ and of the corresponding ranked sample $S_r$, see the following assertions:  %in Figure~\ref{fig:assert_graded_ranked}.
\begin{lstlisting}
len(Sg['rank']) == Sg['K']
len(Sg['inK']) == Sg['K']
sum(Sg['inK']) <= Sg['Np']
Sg['K'] in Sr['K']
Sg['rank'] == Sr['rank'][:Sg['K']]
\end{lstlisting}

\subsubsection{Human-graded samples}\label{sapp:samples_human}
This part consists of 96 samples taken from the Graded Samples (Appendix~\ref{sapp:samples_graded}) and graded by humans. This part was used for estimating Krippendorff coefficients (Section~\ref{ssec:human_scoring}. Each sample has already introduced data fields ($id$, $E$, $N_c$, $N_p$, $K$, $q$, $answer\_ideal$, $answer\_topK$, $grade$) and the human scores: $grade\_human\_i$, where $i$=1,2,3 corresponds to Id of the human.

\subsubsection{Dataset content}\label{sapp:dataset_content}
Each subset of the retrieval-response dataset has comparable amounts of samples with a different number of candidates $N_c$, total number of positives $N_p$ and choice of $K$, subject to the availability in the sources we used. Table~\ref{tab:dataset_content} gives a summary of the dataset content: the number of ranked samples for each embedding, subset, $N_c$, $N_p$ and a range $K_r$ of the selection (``top-K'') choices $K$. The embedding model is not specified, because the numbers of samples are the same for each embedding.

\begin{table}[th!]
%\small
\centering
\begin{tabular}{@{}crrlr@{}}
\hline
{subset}&{$N_c$}&{$N_p$}&{$K_r$}&{$n_r$}\\
\hline
    \parbox[t]{2mm}{\multirow{10}{*}{{A}}}&50&2&[2,15]&300\\
    {}&50&3&[2,15]&300\\
    {}&50&4&[2,15]&300\\
    {}&50&5&[2,15]&300\\
    {}&50&10&[2,15]&300\\
    {}&50&15&[2,20]&300\\
    {}&100&5&[2,15]&300\\
    {}&100&10&[2,15]&300\\
    {}&100&15&[2,20]&300\\
\hline
    \parbox[t]{2mm}{\multirow{1}{*}{{Hp}}}&10&2&[2,5]&900\\
\hline
    \parbox[t]{2mm}{\multirow{7}{*}{{Hs}}}&10&2&[2,5]&900\\
    {}&30&5&[2,15]&719\\
    {}&50&2&[2,15]&900\\
    {}&50&3&[2,15]&900\\
    {}&50&4&[2,15]&803\\
    {}&50&5&[2,15]&298\\
    {}&30&6&[2,15]&231\\
\hline
    \parbox[t]{2mm}{\multirow{8}{*}{{M}}}&30&2&[2,15]&300\\
    {}&30&3&[2,15]&300\\
    {}&30&4&[2,15]&300\\
    {}&30&5&[2,15]&62\\
    {}&50&2&[2,15]&300\\
    {}&50&3&[2,15]&300\\
    {}&50&4&[2,15]&199\\
    {}&50&5&[2,15]&36\\
\hline
    \parbox[t]{2mm}{\multirow{2}{*}{{N}}}&20&2&[2,10]&300\\
    {}&30&2&[2,15]&300\\
\hline
\end{tabular}
\caption{Number $n_r$ of ranked samples for each embedding, subset, $N_c$, $N_p$ and a range of $K$. The number of graded samples equals to $n_r$ multiplied by the number of different $K$ in the range $K_r$. For example, the number of the graded samples in the first row equals 300*(15-2+1)=4200.}
\label{tab:dataset_content}
\end{table}

Table~\ref{tab:dataset_subsets} shows how the samples are split between the narrow ($K<N_p$) and wide ($K>=N_p$) parts of the subsets.

\begin{table}[]
    \centering
    \begin{tabular}{crrrr}
        \toprule
        subset & $n_r$ & $n_g$ & $n_{gn}$ & $n_{gw}$ \\
        \hline
        A & 2700 & 40800 & 15300 & 25500 \\ 
        \hline
        Hp & 900 & 3600 & 0 & 3600 \\
        \hline
        Hs & 4751 & 57514 & 6481 & 51033 \\
        \hline
        M & 1797 & 25158 & 1892 & 23266 \\
        \hline
        N & 600 & 6900 & 0 & 6900 \\
        \bottomrule
    \end{tabular}
    \caption{For each embedding and subset, the counts: Number $n_r$ of ranked samples; number $n_g$ of graded samples; number $n_{gn}$ of narrow graded samples ($K<N_p$); number $n_{gw}$ of wide graded samples ($K>=N_p$).}
    \label{tab:dataset_subsets}
\end{table}

\subsection{Response generation}\label{sapp:prompts_generation}
The following prompt is used during the \textit{Response Generation} phase. 
Given a query and a set of reference documents, the model must generate a response based only on the provided references, without introducing external knowledge.

\noindent\textbf{System Message}
\begin{tcolorbox}[colback=gray!5,colframe=gray!40,breakable]
\small

You are an AI assistant that uses reference documents to respond to a given query.
\end{tcolorbox}

\noindent\textbf{User Message}
\begin{tcolorbox}[colback=gray!5,colframe=gray!40,breakable]
\small

Please respond to the following query according to the information provided in the reference documents. 
Be sure to only use what is in the reference documents to respond to the query and nothing else.

\textbf{Query:}\\
\texttt{<QUERY>}

\textbf{Reference documents:}\\
\texttt{<REFERENCES>}
\end{tcolorbox}

\subsection{Quality score generation}\label{sapp:prompts_score}

In the \textit{Quality Score Generation} phase, an LLM evaluator compares a generated response to its ideal reference response. The model returns a discrete score from 1–5, following the rubric below, to measure content completeness and alignment.

\noindent\textbf{System Message}
\begin{tcolorbox}[colback=gray!5,colframe=gray!40,breakable]
\small

You are an AI assistant who compares a response to its ideal response. 
Given a query, a response, and an ideal response, determine how close the response 
is to the ideal response. Return only a single digit (1, 2, 3, 4, or 5) with no explanations.
\end{tcolorbox}

\noindent\textbf{User Message}
%\begin{tcolorbox}[colback=gray!5,colframe=gray!40,breakable]
\begin{tcolorbox}[colback=gray!5,colframe=gray!40,breakable]
\textbf{RUBRIC}
\small

1 – The response includes substantially less of the relevant information than the ideal response.\\
2 – The response includes about half of the relevant information present in the ideal response.\\
3 – The response includes most of the relevant information present in the ideal response.\\
4 – The response includes nearly all relevant information present in the ideal response.\\
5 – The response includes all relevant information present in the ideal response.\\

\textbf{Query:} [query]\\
\textbf{Response:} [response]\\
\textbf{Ideal Response:} [ideal response]
\end{tcolorbox}

\subsection{LLM vs Human quality scores}\label{sapp:human_scoring}
In Section~\ref{ssec:human_scoring} the LLM grade was compared with human grading by Krippendorff alpha (ordinal) coefficient. Here in Table~\ref{tab:human_llm_spearman} we show Spearman correlations, on the same 96 human scored samples.

\begin{table}[th!]
%\small
\centering
\begin{tabular}{@{}c|rrrr@{}}
\hline
{}&{LLM}&{H1}&{H2}&{H3}\\
\hline
	LLM&&0.83&0.83&0.83\\
	H1&0.83&&0.90&0.81\\
	H2&0.83&0.90&&0.79\\
	H3&0.83&0.81&0.79&\\
\hline
\end{tabular}
\caption{Spearman correlations between the LLM score and three human scores (H1, H2, H3), taken on 96 samples.}
\label{tab:human_llm_spearman}
\end{table}

\section{Response Score}\label{app:response_score}
In Section~\ref{ssec:response_score} we have shown the distribution of the response score for ARXIV, MSMARCO and HotpotQA-sentences. Here we show the distributions for Natural Questions (Figure~\ref{fig:Grade_DatN}) and HotpotQA-paragraphs (Figure~\ref{fig:Grade_DatHp}). They show the same simple pattern as HotpotQA-sentences: the larger the ratio $\frac{K}{N_p}$, the better the response score, meaning that the texts are simple enough for LLM and negative samples are not as important as catching more positives within the range we consider.

Both for generating (Appendix~\ref{sapp:prompts_generation}) and scoring (Appendix~\ref{sapp:prompts_score}) the responses we used LLM GPT-4o-mini.

\section{Correlations with response score}\label{app:Correlations}
For convenience a script is available\footnote{https://github.com/PrimerAI/primer-research}
\subsection{Correlations max by $\alpha$}\label{sapp:corr_max_alpha}
The measures $F$ and $T$ (Equations~\ref{eq:Fa_RP} and~\ref{eq:T}) depend on choice of the parameter $\alpha$. As we noted in Section~\ref{ssec:judgment}, for a fair comparison of measures, for each measure and each data segment we consider a maximal value of a correlation (between a measure and the response score), as achieved over a range of the parameter $\alpha$. 

Through the paper we presented results for data segments defined by the embedding, the dataset and the condition between $K$ and $N_p$ (Section~\ref{ssec:correlations}). A crude condition is a split on samples with $K<N_p$ vs $K\geq N_p$. A more refined condition is by the ratio $\frac{K}{N_p}$, rounded to the first digit.

The search for maximal correlation is done over the grid of 702 values of $\alpha$:
\begin{enumerate}[topsep=0pt,itemsep=-1ex,partopsep=1ex,parsep=1ex]
    \item The edge values $0$ and $1$.
    \item 350 values obtained by raising 0.95 to the power from [1, 2, ..., 350]. 
    \item 350 values obtained by subtracting the above from 1.
\end{enumerate}

\subsection{Narrow and wide subsets}\label{sapp:corr_narrow_wide}
In Section~\ref{ssec:correlations} the Spearman correlations of the measures $T$, $nDCG$ and $F$ are shown in Figure~\ref{fig:SEGcs_comp3_mQ_mNDCG_mF_corr_s}. Here we show the corresponding Pearson and Kendall Tau correlations in Figures~\ref{fig:SEGcs_comp3_mQ_mNDCG_mF_corr_p}, \ref{fig:SEGcs_comp3_mQ_mNDCG_mF_corr_kb}, \ref{fig:SEGcs_comp3_mQ_mNDCG_mF_corr_kc}.

\begin{figure*}[t!]
    \centering
    \includegraphics[width=\textwidth]{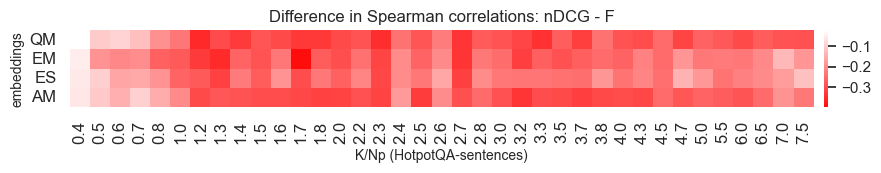} % [width=0.5\textwidth]
    \caption{Difference between the Spearman correlations: $nDCG$-response minus $F$-response. On HotpotQA-sentences.}
    \label{fig:SEGKtoNp_diff_mF_mNDCG_corr_s_DatHs}
\end{figure*}
\begin{figure}[t!]
    \centering
    \includegraphics[width=0.5\textwidth]{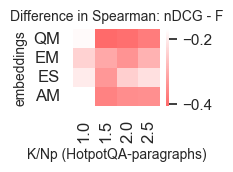} % [width=0.5\textwidth]
    \caption{Difference between the Spearman correlations: $nDCG$-response minus $F$-response. On HotpotQA-paragraphs.}
    \label{fig:SEGKtoNp_diff_mF_mNDCG_corr_s_DatHp}
\end{figure}

\subsection{Subsets by $\frac{K}{N_p}$}\label{sapp:corr_KtoNp}
In Section~\ref{ssec:correlations} the Spearman correlations of $F$ with the response score on ARXIV data were shown in Figure~\ref{fig:SEGKtoNp_corr_mF_corr_s_DatA}. Here in Figures~\ref{fig:SEGKtoNp_corr_mF_corr_s_DatM}, \ref{fig:SEGKtoNp_corr_mF_corr_s_DatN}, \ref{fig:SEGKtoNp_corr_mF_corr_s_DatHs} and \ref{fig:SEGKtoNp_corr_mF_corr_s_DatHp} we show the correlations on the other datasets; this level of Spearman correlations is usually considered as moderate.

Figures~\ref{fig:SEGKtoNp_corr_mQ_corr_s_DatA}, ~\ref{fig:SEGKtoNp_corr_mQ_corr_s_DatM}, \ref{fig:SEGKtoNp_corr_mQ_corr_s_DatN}, \ref{fig:SEGKtoNp_corr_mQ_corr_s_DatHs} and \ref{fig:SEGKtoNp_corr_mQ_corr_s_DatHp} show that correlations of $T$ with the response score are similar across the datasets and the ratios $\frac{K}{N_p}$.

\section{Differences between measures}\label{app:Differences}
In this Section we show more heatmaps illustrating the differences between the measures - for the comparisons that were not decisive enough to be put in the main body of the paper.

Comparison of $nDCG$ and $F$ for measure-response correlations on HotpotQA is shown in Figures~\ref{fig:SEGKtoNp_diff_mF_mNDCG_corr_s_DatHs} and~\ref{fig:SEGKtoNp_diff_mF_mNDCG_corr_s_DatHp}.

\begin{figure}[t!]
    \centering
    \includegraphics[width=0.5\textwidth]{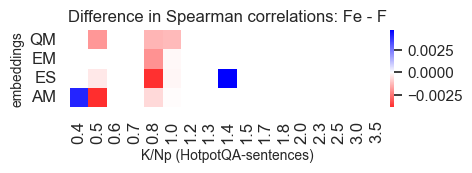} % [width=0.5\textwidth]
    \caption{Difference between the Spearman correlations: $F_e$-response minus $F$-response. On HotpotQA-sentences.}
    \label{fig:EGKtoNp_diff_mF_mE_corr_s_DatHs}
\end{figure}

\begin{figure}[t!]
    \centering
    \includegraphics[width=0.5\textwidth]{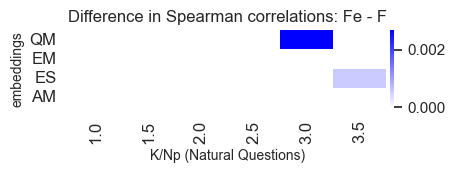} % [width=0.5\textwidth]
    \caption{Difference between the Spearman correlations: $F_e$-response minus $F$-response. On Natural Questions.}
    \label{fig:EGKtoNp_diff_mF_mE_corr_s_DatN}
\end{figure}

Comparison of $F_e$ and $F$ on HotpotQA-sentences is shown in Figure~\ref{fig:EGKtoNp_diff_mF_mE_corr_s_DatHs}, and on Natural Questions in Figure~\ref{fig:EGKtoNp_diff_mF_mE_corr_s_DatN}. There is no noticeable difference on HotpotQA-paragraphs (not shown).

\begin{figure}[t!]
    \centering
    \includegraphics[width=0.4\textwidth]{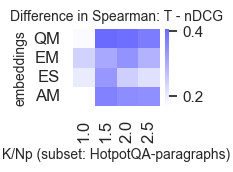} % [width=0.5\textwidth]
    \caption{Difference between the Spearman correlations: $T$-response minus $nDCG$-response. On HotpotQA-paragraphs.}
    \label{fig:SEGKtoNp_diff_mNDCG_mQ_corr_s_DatHp}
\end{figure}

\begin{figure*}[t!]
    \centering
    \includegraphics[width=\textwidth]{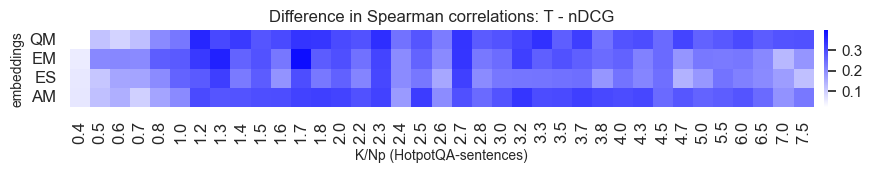} % [width=0.5\textwidth]
    \caption{Difference between the Spearman correlations: $T$-response minus $nDCG$-response. On HotpotQA-sentences.}
    \label{fig:SEGKtoNp_diff_mNDCG_mQ_corr_s_DatHs}
\end{figure*}

\begin{figure*}[t!]
    \centering
    \includegraphics[width=\textwidth]{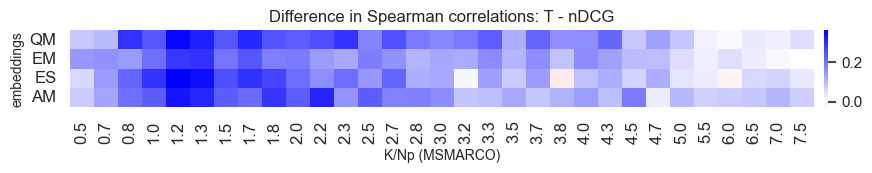} % [width=0.5\textwidth]
    \caption{Difference between the Spearman correlations: $T$-response minus $nDCG$-response. On MSMARCO.}
    \label{fig:SEGKtoNp_diff_mNDCG_mQ_corr_s_DatM}
\end{figure*}

\begin{figure*}[t!]
    \centering
    \includegraphics[width=\textwidth]{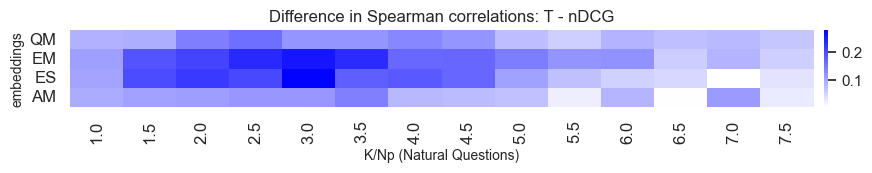} % [width=0.5\textwidth]
    \caption{Difference between the Spearman correlations: $T$-response minus $nDCG$-response. On Natural Questions.}
    \label{fig:SEGKtoNp_diff_mNDCG_mQ_corr_s_DatN}
\end{figure*}

Comparisons between $T$ and $nDCG$ was shown for ARXIV in Figure~\ref{fig:SEGKtoNp_diff_mNDCG_mQ_corr_s_DatA}. Here it is shown for other datasets in Figures~\ref{fig:SEGKtoNp_diff_mNDCG_mQ_corr_s_DatHp}, \ref{fig:SEGKtoNp_diff_mNDCG_mQ_corr_s_DatHs}, \ref{fig:SEGKtoNp_diff_mNDCG_mQ_corr_s_DatM}, \ref{fig:SEGKtoNp_diff_mNDCG_mQ_corr_s_DatN}.

\end{document}